\documentclass{article}

\usepackage[preprint]{neurips_2025}

\usepackage[utf8]{inputenc} 
\DeclareUnicodeCharacter{2212}{\textminus}
\usepackage[T1]{fontenc}    
\usepackage{hyperref}       
\usepackage{url}            
\usepackage{booktabs}       
\usepackage{amsfonts}       
\usepackage{nicefrac}       
\usepackage{microtype}      
\usepackage{xcolor}         
\usepackage{amsmath,amssymb,amsthm}
\usepackage[nameinlink,noabbrev]{cleveref} 
\usepackage{enumitem} 
\usepackage{multicol}
\usepackage{graphicx}
\usepackage{wrapfig}

\newtheorem{theorem}{Theorem}[section]

\theoremstyle{definition}

\newtheorem{assumption}[theorem]{Assumption}

\theoremstyle{remark}

\crefname{assumption}{Assumption}{Assumptions}
\crefname{theorem}{Theorem}{Theorems}
\usepackage{multirow}

\title{Graph Contrastive Learning via Spectral Graph Alignment}

\author{%
  Manh Nguyen \\
  Department of Statistics\\
University of Wisconsin-Madison \\
  \texttt{mdnguyen4@wisc.edu} \\
}

\begin{document}

\maketitle

\begin{abstract}
Given augmented views of each input graph, contrastive learning methods (e.g., InfoNCE) optimize pairwise alignment of graph embeddings across views while providing no mechanism to control the global structure of the view specific graph-of-graphs built from these embeddings. We introduce \textbf{SpecMatch-CL}, a novel loss function that aligns the view specific graph-of-graphs by minimizing the difference between their normalized Laplacians. Theoretically, we show that under certain assumptions, the difference between normalized Laplacians provides an upper bound not only for the difference between the ideal Perfect Alignment contrastive loss and the current loss, but also for the Uniformly loss \cite{wang2020understanding}. Empirically, SpecMatch-CL establishes new state of the art on eight TU benchmarks under unsupervised learning and semi-supervised learning at low label rates, and yields consistent gains in transfer learning on PPI-306K and ZINC 2M datasets. The Pytorch implementation for our method is provided in \href{https://github.com/manhbeo/GNN-CL}{github.com/manhbeo/GNN-CL}.

\end{abstract}

\section{Introduction}
Contrastive learning (CL) has become a dominant paradigm for representation learning across modalities, including graphs \cite{chen2020simple,you2020graphcl,chu2021coco}. At its core, CL encourages the alignment of positive pairs while maintaining uniformity in the representation space \cite{wang2020understanding}. In graphs, recent approaches realize these principles either by manipulating topology through augmentations (e.g. GraphCL and its automated variants JOAO/JOAOv2) or by contrasting local/global summaries and diffusion-based views \cite{you2020graphcl,you2021joao,you2021joaov2,sun2020infograph,hassani2020mvgrl}. However, the instance-level nature of InfoNCE leaves an important degree of freedom: two augmented views can attain similar contrastive objectives while inducing different global neighborhood structure (e.g., connectivity patterns, cluster margins, and multihop relations).

We address this gap by explicitly regularizing view-to-view spectral consistency. Our method, \textbf{SpecMatch-CL}, constructs sparse neighborhood graphs from the embeddings of each view and penalizes the spectral norm of the difference between their normalized Laplacians. Intuitively, if the spectra match, the induced diffusion geometries --and hence multiscale neighborhoods---are similar across views, removing degeneracies that alignment alone does not fix. To ground this design, we show that small Laplacian discrepancy implies not only a small discrepancy of the contrastive objective relative to its ideal Perfect Alignment counterpart but also a small Uniformity loss, thereby providing an optimization-agnostic rationale for the novel loss.

\paragraph{Contributions:}  
\begin{itemize}
    \item[1)] We propose \textbf{SpecMatch-CL}, a spectral graph-matching regularizer that aligns the normalized Laplacian of view-wise embedding graphs and integrates seamlessly with GraphCL-style training.
    \item[2)] We develop an analysis showing that the spectral loss $\mathcal{L}_G$ simultaneously controls (i) the gap between contrastive loss and its Perfect Alignment counterpart and (ii) the Wang–Isola uniformity objective, which explains how enforcing global spectral alignment benefits both alignment and uniformity in instance-level contrastive learning.
    \item[3)] We demonstrate consistent improvements in graph classification under unsupervised and semi-supervised regimes and in transfer to molecular / biological property prediction, achieving state-of-the-art results while keeping the training recipe and enhancements unchanged.
\end{itemize}

\section{Related work}
\subsection{Contrastive learning on graphs}

Self-supervised learning on graphs has been largely driven by contrastive objectives that adapt InfoNCE-style losses to node- and graph-level tasks. Early methods such as Deep Graph Infomax (DGI) maximize mutual information between local node representations and a global graph summary, producing strong node-level embeddings without labels \cite{velickovic2019dgi}. Subsequent work has explored view generation and augmentation at scale. GraphCL adapts InfoNCE loss to the graph domain by applying hand-crafted graph augmentations (node drop, edge perturbation, attribute masking, subgraph sampling) and contrasting two augmented views of each graph, showing that simple GNN backbones with appropriate augmentations already yield strong unsupervised and transfer performance \cite{you2020graphcl}. GRACE and its adaptive variant GCA extend contrastive learning to node-level pretraining through structural and feature corruptions \cite{zhu2020grace,zhu2021gca}. JOAO also automates the selection of augmentations for GraphCL through a bi-level optimization scheme that chooses augmentations per data set and training step \cite{you2021joao}. Across these methods, the primary focus is on instance-level alignment: embeddings of the same node or graph under two augmentations are pulled together, while other examples in the batch serve as negatives. The global geometry of the embedding space---and, in particular, the induced similarity graph over all graphs within a view---remains largely unconstrained.

Recent theoretical work has begun to reinterpret such contrastive objectives in spectral terms. Tan et al.\ prove that, under standard design choices (normalized embeddings and a Gaussian kernel), minimizing InfoNCE is equivalent to performing spectral clustering on a similarity graph whose edge weights encode positive-pair sampling probabilities \cite{tan2024spectral}. Complementarily, HaoChen et al.\ show that minimizing the Spectral contrast Loss (SCL), defined on an augmentation graph whose nodes are augmented views and whose edges connect augmentations of the same underlying example, effectively performs a spectral decomposition of this augmentation graph and produces representations with provable linear-probe guaranties \cite{haochen2021scl}. Inspired by these spectral perspectives, our work transfers the idea from an augmentation graph over individual examples to a graph-of-graphs built from graph-level embeddings within each view: we explicitly regularize the normalized Laplacian of the view-specific similarity graphs, rather than relying on InfoNCE alone to shape their global structure.

\subsection{Alignment and Uniformity}

Wang and Isola analyze InfoNCE through two geometric functionals in the feature distribution: \emph{Alignment} and \emph{Uniformity} in the unit hypersphere \cite{wang2020understanding}. Specifically, denote $z_i, z_j\in\mathbb{R}^d$ as the normalized embeddings ($z_i$ and $z_j$ are in the unit sphere $\mathbb{S}^{d-1}$), and let $p_{\text{pos}}(\cdot)$ and $p_{\text{data}}(\cdot)$ denote the distribution of positive embeddings in $\mathbb{R}^d\times \mathbb{R}^d$ and the embedding distribution in $\mathbb{R}^d$, respectively. The Alignment loss measures the expected distance between embeddings of positive pairs:
\begin{align}
    \mathcal{L}_{\mathrm{align}}
    \;=\;
    \mathbb{E}_{(z_i, z_j)\sim p_{\text{pos}}}
    \big\| z_i -  z_j \big\|_2^{\alpha},
    \qquad \alpha > 0,
\end{align}
and is small when augmentations of the same instance are mapped to nearby points. Uniformity instead quantifies how well the embedding distribution spreads out on the hypersphere via a Gaussian (RBF) potential:
\begin{align}
    \mathcal{L}_{\mathrm{unif}} 
= \log \, \mathbb{E}_{\substack{z_i, z_j \overset{\mathrm{iid}}{\sim} p_{\text{data}}}}
\left[
    e^{-t \| z_i - z_j \|_2^2}
\right]
\end{align}
which is minimized when the embeddings are approximately uniformly distributed in $\mathbb{S}^{d-1}$. Wang and Isola show that, in the limit of many negatives and appropriately chosen temperature, the standard contrastive loss asymptotically optimizes a trade-off between decreasing $\mathcal{L}_{\mathrm{align}}$ (tight clusters for positives) and decreasing $\mathcal{L}_{\mathrm{unif}}$ (globally repulsive, nearly uniform configurations) \cite{wang2020understanding}.

Within this framework, they further formalize two idealized regimes. An encoder $f$ is said to achieve \emph{Perfect Alignment} if $z_i = z_j$ almost surely (a.s.) for $(z_i, z_j)\sim p_{\text{pos}}$, i.e., all augmentations of the same instance collapse to a single point on the hypersphere, and it achieves Perfect Uniformity if the distribution of embeddings $p_{\text{data}}(\cdot)$ is the uniform distribution on the unit sphere $\mathbb{S}^{d-1}$. Recent contrastive methods, therefore, minimize both loss to achieve desired performance. Our analysis further shows that reducing the novel spectral graph matching loss $\mathcal{L}_{G}$ not only tightens an upper bound on the contrastive loss gap to Perfect Alignment but also upper-bounds the Uniformity loss.

\section{Method}
\begin{figure}
    \centering
    \includegraphics[width=1\linewidth]{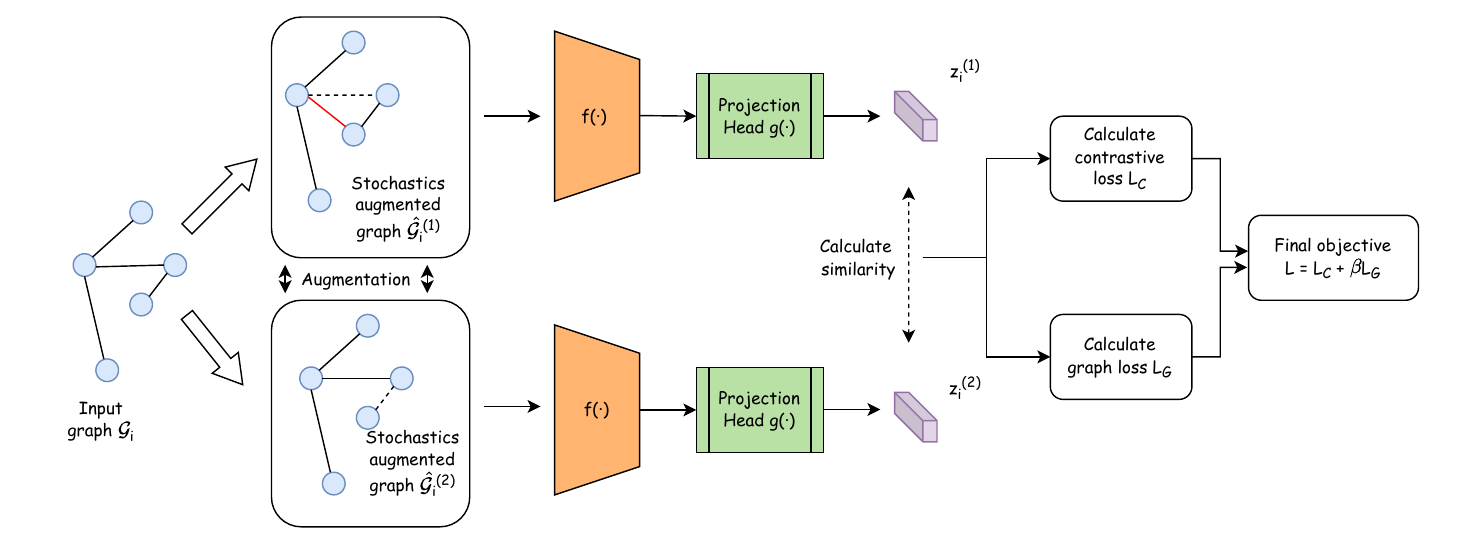}
    \caption{An illustration of SpecMatch-CL method. }
    \label{fig:methods}
\end{figure}

\subsection{Problem definition}
In this paper, we focus on graph-level contrastive learning. Let $\mathcal{G}_i = (\mathcal{V}_i, \mathcal{E}_i)$ be an undirected graph, where $\mathcal{V}_i = \{v_k\}_{k=1}^{|\mathcal{V}_i|}$ is the set of nodes and $\mathcal{E} = [e_{kj}] \in \mathbb{R}^{|\mathcal{V}_i|\times|\mathcal{V}_i|}$ is the adjacency matrix. Each node $v_n$ is associated with an attribute vector $\textbf{x}_n \in \mathbb{R}^{N}$, and we collect them into a feature matrix $\textbf{X} \in \mathbb{R}^{|\mathcal{V}_i|\times N}$ with $\textbf{x}_n = \textbf{X}[n,:]^{\top}$. In the graph-level setting, we are given a collection of unlabeled graphs
\[
\mathcal{G} = \{\mathcal{G}_1, \mathcal{G}_2, \dots, \mathcal{G}_N\},
\]
and the goal is to learn an encoder that maps each graph $\mathcal{G}_i \in \mathcal{G}$ to a $d$-dimensional representation $z_{i} \in \mathbb{R}^d$ using only the structure and features of the graphs.

\subsection{Graph augmentations}
Following the standard two-view contrastive setup in You et al.\cite{you2020graphcl}, for a graph $\mathcal{G}_i \in \mathcal{G}$ we sample two augmented views
\[
\hat{\mathcal{G}}_i^{(1)},\hat{\mathcal{G}}_i^{(2)} \sim \mathcal{T}(\,\cdot \mid \mathcal{G}_i\,),
\]
where $\mathcal{T}(\,\cdot \mid \mathcal{G}_i\,)$ is an augmentation distribution conditioned on $\mathcal{G}_i$, encoding prior assumptions about plausible perturbations of the graph. We consider four basic augmentation operators for constructing positive pairs of graphs:
\begin{enumerate}
    \item \emph{Node dropping}: randomly remove a subset of nodes together with their incident edges;
    \item \emph{Edge perturbation}: randomly add or delete a subset of edges to alter local connectivity;
    \item \emph{Attribute masking}: hide a subset of node features and require the encoder to reconstruct or ignore the missing information from context;
    \item \emph{Subgraph sampling}: select a subgraph of $\mathcal{G}_i$ using, e.g., a random walk procedure.
\end{enumerate}

\subsection{Graph encoder}
Once the augmented graph pair $(\hat{\mathcal{G}}_i^{(1)}, \hat{\mathcal{G}}_i^{(2)})$ is generated, we feed each view into a graph encoder to obtain their representations. Our framework is agnostic to the specific architecture, but throughout we use a graph neural network (GNN) to encode graphs. Specifically, for a view $v\in\{1,2\}$, let $\hat{\mathcal{G}}_i^{(v)} = (\mathcal{V}_i^{(v)}, \mathcal{E}_i^{(v)})$ with node features $\textbf{X} \in \mathbb{R}^{|\mathcal{V}_i^{(v)}|\times N}$, where $\textbf{x}_n = \textbf{X}[n,:]^\top$ is the feature vector of node $v_n \in \mathcal{V}_i^{(v)}$. Consider an $L$-layer GNN $f(\cdot)$. The message-passing update at the $l$-th layer is
\begin{align}
\textbf{a}_n^{(l)} &= \mathrm{AGGREGATION}^{(l)}\big(\{\textbf{h}_{n'}^{(l-1)} : n' \in \mathcal{N}(n)\}\big), \\
\textbf{h}_n^{(l)} &= \mathrm{COMBINE}^{(l)}\big(\textbf{h}_n^{(l-1)}, \textbf{a}_n^{(l)}\big),
\end{align}
where $\textbf{h}_n^{(l)}$ is the embedding of node $v_n$ at layer $l$, initialized with $\textbf{h}_n^{(0)} = \textbf{x}_n$, $\mathcal{N}(n)$ denotes set of nodes adjacent to $v_n$, and $\mathrm{AGGREGATION}^{(l)}(\cdot)$ and $\mathrm{COMBINE}^{(l)}(\cdot)$ are layer-specific functions (e.g., mean/sum aggregation and nonlinear transformation).  

After $L$ layers of propagation, node embeddings are pooled into a graph-level representation via a READOUT function (e.g., sum/mean pooling or attention), then a multi-layer perceptron $g(\cdot)$ is used for downstream graph-level tasks, and the same encoder is shared for all augmented views in the contrastive learning framework.
:
\begin{align*}
    &f(\hat{\mathcal{G}}_i^{(v)})
= \text{READOUT}\bigl(\{ \mathbf{h}_{n'}^{(l-1)} : v_n \in \mathcal{V}_i^{(v)},\, l \in L \}\bigr),\\
&z_i^{(v)} = g\left(f(\hat{\mathcal{G}}_i^{(v)})\right).
\end{align*}

\subsection{Contrastive Learning Framework}
For the view $v\in \{1,2\}$, define the other view as $v'$. As shown by Chen et al. \cite{chen2020simple} and Khosla et al. \cite{khosla2020supervised}, normalizing the embeddings helps increase the performance 
\begin{align*}
    z^{(v)}_i \leftarrow \frac{z^{(v)}_i}{\|z^{(v)}_i\|}, \qquad z^{(v')}_i \leftarrow\ \frac{z^{(v')}_i}{\|z^{(v')}_i\|}
\end{align*}
For the normalized embeddings, we define the similarity as inner product
\begin{align*}
    s(z^{(v)}_i,z^{(v')}_i) = z^{(v)\top}_iz^{(v')}_i
\end{align*}
Following the previous works of graph contrastive learning \cite{you2020graphcl} \cite{chu2021coco} \cite{liu2023mssgcl}, we use normalized temperature-scaled cross-entropy (InfoNCE) loss \cite{oord2018cpc} as the contrastive objective. The single sample contrastive loss according to view $v$ can be written as
\[
l_{i}^{(v)}
= -\log \frac{\exp(s(z^{(v)}_i, z^{(v')}_i)/\tau)}
{\sum_{a=1}^2\sum_{k=1}^{N} \exp\!\big(s(z^{(v)}_i, z^{(a)}_k)/\tau\big) \cdot (1 - \mathbf{1}\{a=v\wedge k=i\})}
\]
where $\mathbf{1}(\cdot)$ is the indicator function and $\tau$ is the temperature. We then have the total contrastive objective as
\[
\mathcal{L}_{C} \;=\; \sum_{v=1}^2 \sum_{i=1}^{N} l_{i}^{(v)}.
\]

\subsection{Spectral Graph Matching}
Given embedding vectors from two views of the data, we optimize the spectral graph matching algorithm by constructs corresponding graphs and compares their spectral properties to ensure structural consistency. Particularly, we construct the similarity matrix $S^{(v)}$ whose entries are defined as: 
\begin{align*}
    S_{ij}^{(v)} &= s(z_i^{(v)}, z_j^{(v)})
\end{align*}
The adjacency matrices $A^{(v)}$ are then formed by thresholding the similarities:
\begin{align*}
    A_{ij}^{(v)} &= \begin{cases}
        1, & \text{if } S_{ij}^{(v)} > \theta \text{ and } i \neq j \\
        0, & \text{otherwise}
    \end{cases}
\end{align*}
with the similarity threshold $\theta$. An example of the adjacency matrices is shown in Figure 2. 
\begin{wrapfigure}{r}{0.5\textwidth}
    \includegraphics[width=1\linewidth]{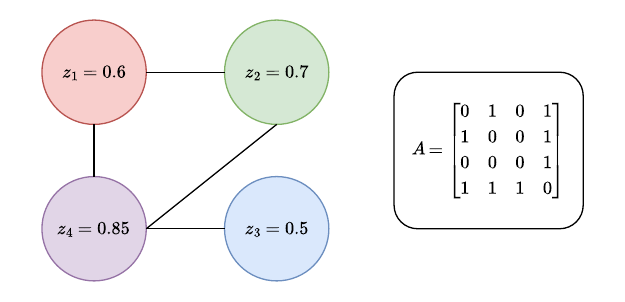}
    \caption{A 1D example of the adjacency matrix $A$ with $\theta = 0.4$.}
    \label{fig:placeholder}
\end{wrapfigure} Instead of using a fixed threshold, we determine the threshold adaptively based on the distribution of similarity values:
\begin{align*}
    \theta = Q(S^{(v)}, p),
\end{align*}
where $Q(S^{(v)}, p)$ is the $p$-th percentile of the similarity values in $S^{(v)}$. From each adjacency matrix, we compute the corresponding degree matrices $D^{(v)}$ as 
\begin{align*}
    D^{(v)}_{ii} = \sum_{j} A^{(v)}_{ij}
\end{align*}
The normalized Laplacian matrices are defined as:
\begin{align*}
    L^{(v)} &= I - \left(D^{(v)}\right)^{-\frac{1}{2}} A^{(v)} \left(D^{(v)}\right)^{-\frac{1}{2}},
\end{align*}
and the spectral graph matching loss is then computed by:
\begin{align*}
    \mathcal{L}_{G} = ||L^{(1)} - L^{(2)}||_F^2.
\end{align*}

\subsection{SpecMatch-CL loss}
The total SpecMatch-CL loss combines the contrastive loss and the spectral graph alignment terms:
\begin{align*}
    \mathcal{L} = \mathcal{L}_{C} + \beta \mathcal{L}_{G},
\end{align*}
where $\alpha$ balances the contributions of each loss component. Figure 1 provides an illustration for the process of our method.

\section{Theoretical Justification}
To theoretically justify SpecMatch-CL, we derive two theorems, built on the Alignment and Uniformity losses introduced by Wang and Isola \cite{wang2020understanding}, showing how the spectral graph-matching loss $\mathcal{L}_{G}$ controls both the gap to the Perfect Alignment case and the Uniformity objective.

\subsection{$\mathcal{L}_{G}$ provides an upper bound for the Contrastive loss gap to Perfect Alignment}
In this section, we clarify how the spectral graph-matching loss $\mathcal{L}_{G}$ influences the contrastive objective. Although InfoNCE operates directly on pairwise similarities between embeddings, $\mathcal{L}_{G}$ measures discrepancies between view-wise graph geometries encoded by their normalized Laplacians. Our goal is to relate these two levels by showing that if the diffusion geometries of the two views are close, then the realized contrastive loss cannot deviate far from its ideal Perfect Alignment value.

While $\mathcal{L}_{G}$ penalizes discrepancies between the two view-wise normalized Laplacians, the contrastive objective acts on pairwise similarities between learned embeddings. To connect the two levels of abstraction, we employ the diffusion (heat) kernel \cite{coifman2006diffusionmaps,jones2008manifold} as a geometry-aware operator on the graphs induced by each view. Concretely, with a diffusion scale \(t_d > 0\), we define the heat kernel for each view $v$ as $P^{(v)} := \exp(-t_dL^{(v)})$, then the associated diffusion distance for each input instance is $\left\|P^{(1)} - P^{(2)} \right\|_F^2$. We posit a mild consistency link between the distances between the two embeddings and the distances their induced diffusion geometries drift. Intuitively, if the embeddings of a positive pair are separate, the neighborhoods they activate in the two views should also look different.

\begin{assumption} \textit{We assume that there exist a constant $c$ for input graphs $\mathcal{G}$ such that} 
\begin{align*}
    \sum_{i=1}^N\left\| z^{(1)}_i  - z^{(2)}_i\right\|_2^2 \le c \left\|P^{(1)} - P^{(2)} \right\|_F^2
\end{align*}
\end{assumption}

This implies that diffusion mismatch upper-bounds embedding mismatch up to a data-dependent factor $c$. The assumption is violated in the degenerate case $\left\|P^{(1)} - P^{(2)} \right\|_F^2 = 0$, i.e., when $P^{(1)} = P^{(2)}$ (equivalently $L^{(1)} = L^{(2)}$) and the two view-induced graphs are identical, which is rare in practice. With that assumption, we introduce the theorem: 
\begin{theorem} Under Assumption 4.1 we have
\begin{align*}
    \left|\mathcal{L}_{C} - \mathcal{L}_{C}^{*}\right| &\le \frac{(t_d)^2 c}{\tau} \mathcal{L}_{G}
\end{align*}
a.s. over $(z_i^{(1)}, z_i^{(2)}) \sim p_{pos}$, where $\tau$ is the temperature in contrastive loss and $t_d$ is the diffusion scale.
\end{theorem}
The proof for Theorem 4.2 is provided in Appendix A.
\paragraph{Discussion.}
Under Assumption 4.1, the theorem establishes a link between the spectral alignment and the contrastive objective: a reduction in the spectral graph-matching loss $\mathcal{L}_{G}$ produces a provably tighter upper bound on the deviation between the realized contrastive loss and its Perfect Alignment counterpart. This observation provides a rationale for adding the graph matching loss: by constraining the view-wise Laplacians to be close, not only the geometry of the two views are aligned but also the contrastive objective becomes closer to its ideal value, which is consistent with improvements in unsupervised, semi-supervised and transfer accuracy as shown in the experiment section.

\subsection{$\mathcal{L}_{G}$ provides an upper bound for the Uniformity loss}
Having established that the spectral graph-matching loss $\mathcal{L}_{G}$ controls the deviation from the Perfect Alignment contrastive objective, we now turn to its effect on uniformity. Recall that, in the Wang--Isola framework, uniformity quantifies how well the embedding distribution spreads out on the unit sphere, penalizing collapsed or highly clustered configurations. Our goal here is to show that enforcing a small spectral discrepancy between view-wise Laplacians also drives the encoder toward low uniformity loss, i.e., toward a more evenly dispersed configuration of graph-level embeddings. Intuitively, if the two views induce similar diffusion geometries on the same node set, then their embeddings cannot concentrate in a few narrow regions without incurring a large Laplacian mismatch.

Assume that each augmentation graph is connected; denote $d^{(v)}_i := D^{(v)}_{ii}$ as the $i^{th}$ diagonal elements of $D^{(v)}$ and $\lambda_2$ as the smallest non-zero eigenvalue of the normalized Laplacian matrix $L$. In addition, let the distribution of $L$ (over the randomness of the augmentations) be conditional on the set of input graphs $\mathcal{G}$ as $\mathbb{P}_{L|\mathcal{G}}$. Additionally, denote $\overline{L} := \mathbb{E}_{L\sim \mathbb{P}_{L|\mathcal{G}}}\!\left[L\right]$ and $\overline{\lambda}_2$ as the smallest non-zero eigenvalue of $\overline{L}$. Recall the Wang--Isola uniformity potential at temperature $t > 0$:
\[
\mathcal{L}_{\mathrm{unif}} 
= \log \, \mathbb{E}_{\substack{z_i, z_j \overset{\mathrm{iid}}{\sim} p_{\text{data}}}}
\left[
    e^{-t \| z_i - z_j \|_2^2}
\right].
\]
Let $Z$ be the matrix of embeddings used to construct $L$. For embeddings $z_i$ (rows of $Z$), let the degree-weighted mean be 
\begin{align*}
    \mu := \frac{1}{\sum_{k=1}^N d_k} \sum_{i=1}^N d_i z_i.
\end{align*}

\begin{theorem} Assume that each augmentation graph is connected and $L^{(1)}$ and $L^{(2)}$ are i.i.d.\ (conditional on $\mathcal{G}$). We have
\begin{align*}
    \mathcal{L}_{\mathrm{unif}} \le \frac{1 - e^{-4t}}{2\sqrt{2}}\left(\frac{3}{2} - \mathbb{E}\left[\|\mu\|_2^2\right]\right)\ \sqrt{\mathbb{E}_{L^{(1)}, L^{(2)}\sim\mathbb{P}_{L|\mathcal{G}}}[\mathcal{L}_G]}  - \frac{(1 - e^{-4t})}{2}\overline{\lambda}_2(1 - \mathbb{E}\left[\|\mu\|_2^2\right]),
\end{align*}
where $t$ is the temperature parameter of the Uniformity loss.
\end{theorem}
The proof for Theorem 4.3 is provided in Appendix B.

\paragraph{Discussion.}
Theorem~4.3 links our spectral regularizer directly to the Wang--Isola uniformity objective. The first term on the right-hand side shows that $\mathcal{L}_{\mathrm{unif}}$ grows at most like $\sqrt{\mathbb{E}[\mathcal{L}_{G}]}$, so reducing the spectral graph-matching loss tightens a nontrivial upper bound on uniformity: better spectral alignment between views forces the embedding distribution to be more spread out. The term $|\mu|_2^2$ is the squared norm of the degree-weighted mean embedding: when representations are well balanced on the sphere, $|\mu|_2^2$ is small and $(1 - \mathbb{E}\left[|\mu|_2^2\right])$ is close to one, strengthening the negative contribution of the second term; when the encoder collapses the mass into a few directions, $|\mu|_2^2$ grows and the bound weakens. The dependence on $\overline{\lambda}_2$, the smallest non-zero eigenvalue of the expected normalized Laplacian, ties uniformity to graph connectivity: a larger spectral gap (better-connected similarity graph) tightens the bound and promotes more uniform embeddings.

\begin{wrapfigure}{r}{0.6\textwidth}
    \includegraphics[width=1\linewidth]{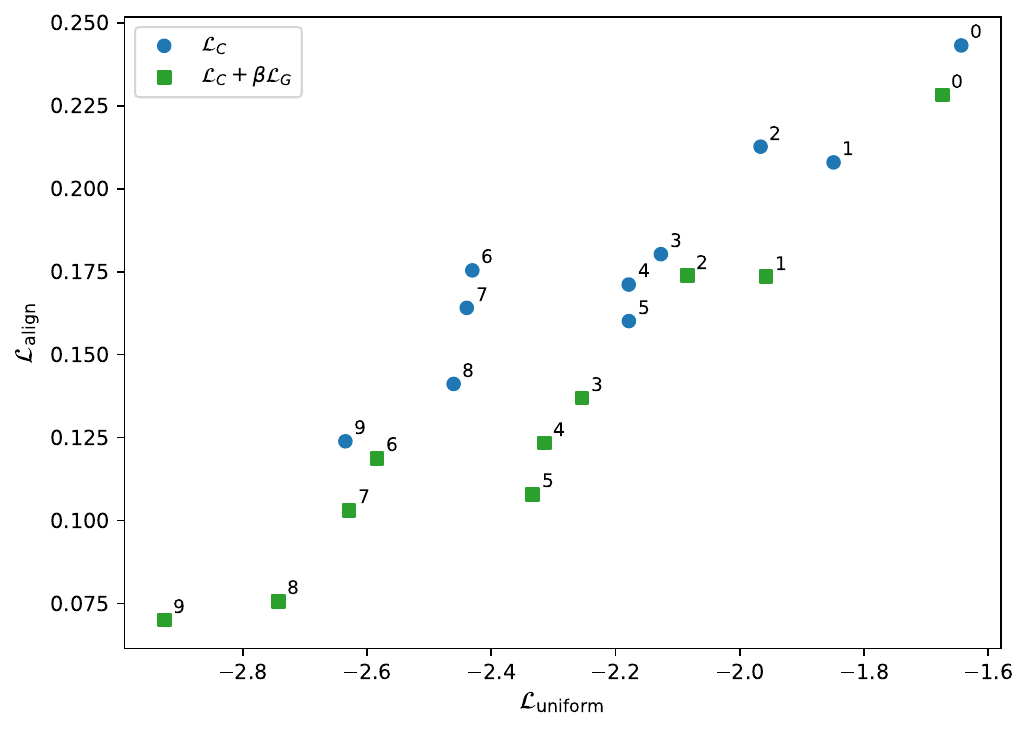}
    \caption{ $\mathcal{L}_{align} - \mathcal{L}_{unif}$ plot for contrastive learning with and without the spectral graph matching loss on NCI1 dataset. The numbers around the points are the indexes of epochs. For both $\mathcal{L}_{align}$ and $\mathcal{L}_{unif}$, lower is better.}
    \label{fig:3}
\end{wrapfigure} We empirically assess the impact of the spectral graph-matching loss by tracking checkpoints of contrastive training with and without this term every two epochs when training on NCI1 dataset and plotting their alignment and uniformity losses in Figure~\ref{fig:3}. In this experiment, we use $\alpha = 2$ for $\mathcal{L}_{\mathrm{align}}$, $t=2$ for $\mathcal{L}_{\mathrm{unif}}$, and $\beta = 0.5$. While both variants progressively improve $\mathcal{L}_{\mathrm{align}}$ and $\mathcal{L}_{\mathrm{unif}}$, the model with the spectral graph-matching loss achieves consistently faster reductions in both metrics. Taken together, the theorems show that minimizing $\mathcal{L}_{G}$ simultaneously pulls the contrastive loss toward its Perfect Alignment limit and acts as a spectral surrogate for the uniformity criterion, providing a unified explanation for the empirical gains observed across unsupervised, semi-supervised, and transfer settings.

\section{Experiments}
We evaluate SpecMatch-CL to test whether spectral graph matching consistently improves over strong contrastive baselines, covering three regimes: unsupervised learning, semi-supervised learning (both are conducted on TU benchmarks), and transfer learning from large-scale pre-training to molecular and biological prediction tasks.

\subsection{Experimental Setup}
\paragraph{Training framework and hyperparameters}
We adopt the GraphCL training framework as the base \cite{you2020graphcl}. We follow its default augmentation strength (0.2) and choose augmentation operators by data regime: for biochemical molecules, we apply node dropping and subgraph extraction; for dense social networks, we use all four operators; and for sparse social networks, we use all except attribute masking. For our adaptive similarity threshold, we set the percentile to $p\!=\!80$ by default (ablation reported in Appendix C). We align our backbones and hyperparameters with widely used settings to ensure comparability across regimes. (1) \emph{Unsupervised representation learning}: we use GIN with 3 layers and 32 hidden dimensions \cite{xu2019how}. (2)  \emph{Semi-supervised learning}: we use a 5-layer ResGCN with 128 hidden dimensions \cite{chen2019dissect}. (3)  \emph{Transfer learning}: we use GIN with 5 layers and 300 hidden dimensions, following standard practice \cite{hu2020ogb}. Unless specified, the weight of our spectral loss is $\beta\in\{0.5,0.75,1.0,1.25,1.5\}$, selected by grid search (for ablation study of $\beta$, see Appendix D). 

\paragraph{Datasets}
For unsupervised and semi-supervised graph-level learning, we follow standard practice and evaluate on the TU benchmark collection \cite{morris2020tudataset}, which comprises social network graphs: COLLAB, IMDB-B/M, RDT-B, RDT-M5K, commonly traced to graph kernel benchmarks \cite{yanardag2015dgk,shervashidze2011wl}, as well as biochemical molecule datasets: MUTAG, NCI1, PROTEINS, DD. For transfer learning, we pre-train on large unlabeled dataset ZINC 2M (from the ZINC database) \cite{sterling2015zinc} and PPI-306K (as in graph pre-training protocols) \cite{hu2020pretrain}—and fine-tune on downstream suites: molecular property prediction tasks from MoleculeNet: BBBP, ToxCast, SIDER, ClinTox, MUV, HIV, BACE \cite{wu2018moleculenet},  and protein–protein interaction classification dataset PPI \cite{hamilton2017graphsage}.

\paragraph{Evaluation protocols}
Following widely adopted evaluation protocols for graph-level self-supervised learning \cite{sun2020infograph,you2020graphcl,you2021joao,you2021joaov2}, we assess generalization in both unsupervised and semi-supervised regimes. In the unsupervised setting, we train SpecMatch-CL on the full training graphs to obtain graph-level embeddings and then fit a downstream linear SVM with 10-fold cross-validation on each dataset \cite{narayanan2017graph2vec}. In the semi-supervised setting, we pre-train GNNs with SpecMatch-CL on all available training graphs and fine-tune with the prescribed label-rate protocol (stratified $K$-fold when explicit splits are unavailable; otherwise train/val/test splits as provided by the benchmark). Hyperparameters for fine-tuning are selected on validation sets; we report mean and standard deviation over multiple random seeds. 

\paragraph{Baselines}
We compare SpecMatch-CL against (i) classical graph-kernel methods: Weisfeiler–Lehman (WL) \cite{shervashidze2011wl} and Deep Graph Kernels (DGK) \cite{yanardag2015dgk}; (ii) unsupervised graph embedding baselines—sub2vec \cite{adhikari2018sub2vec}, node2vec \cite{grover2016node2vec}, and graph2vec \cite{narayanan2017graph2vec}; and (iii) representative graph contrastive/self-supervised methods—InfoGraph \cite{sun2020infograph}, GraphCL \cite{you2020graphcl}, JOAO and JOAOv2 \cite{you2021joao,you2021joaov2}, SimGRACE \cite{xia2022simgrace}, Multi-Scale Subgraph Contrastive Learning (MSSGCL) \cite{liu2023mssgcl}, and CuCo (Curriculum Contrastive Learning) \cite{chu2021coco}. All baselines are trained under their recommended settings to ensure comparability. 

\subsection{Unsupervised training}
As shown in Table~1, under the same GraphCL training recipe and augmentation strength, SpecMatch-CL attains state-of-the-art accuracy on all eight TU benchmarks. Compared with the strongest prior method on each dataset, it improves performance by about \(0.61\) percentage points on average (e.g., \(+0.41\) on NCI1, \(+0.78\) on PROTEINS, \(+1.39\) on DD, \(+0.32\) on MUTAG, \(+0.78\) on COLLAB, \(+0.23\) on RDT-B, \(+0.73\) on RDT-M5K, and \(+0.21\) on IMDB-B). These consistent gains indicate that enforcing view-to-view spectral alignment provides complementary benefits to standard instance-level objectives and augmentation design, leading to uniformly stronger graph-level representations under unsupervised pretraining.

\begin{table*}[t]
\centering
\caption{Graph classification accuracy (\%) on benchmark datasets under unsupervised training. Best results are presented in bold}
\resizebox{\linewidth}{!}{%
\begin{tabular}{l|cccccccc}
\toprule
Method & NCI1 & PROTEINS & DD & MUTAG & COLLAB & RDT-B & RDT-M5K & IMDB-B \\
\midrule
WL            & 80.01$\pm$0.50 & 72.92$\pm$0.56 & 74.02$\pm$2.28 & 80.72$\pm$3.00 & 60.30$\pm$3.44 & 68.82$\pm$0.41 & 46.06$\pm$0.21 & 72.30$\pm$3.44 \\
DGK           & 80.31$\pm$0.46 & 73.30$\pm$0.82 & 74.85$\pm$0.74 & 87.44$\pm$2.72 & 64.66$\pm$0.50 & 78.04$\pm$0.39 & 41.27$\pm$0.18 & 66.96$\pm$0.56 \\
\midrule
sub2vec       & 52.84$\pm$1.47 & 53.03$\pm$5.55 & 54.33$\pm$2.44 & 61.05$\pm$15.80 & 55.26$\pm$1.54 & 71.48$\pm$0.41 & 36.68$\pm$0.42 & 55.26$\pm$1.54 \\
node2vec      & 54.89$\pm$1.61 & 57.49$\pm$3.57 & 74.77$\pm$0.51 & 72.63$\pm$10.20 & 54.57$\pm$0.37 & 72.76$\pm$0.92 & 31.09$\pm$0.14 & 58.02$\pm$2.30 \\
graph2vec     & 73.22$\pm$1.81 & 73.30$\pm$2.05 & 70.32$\pm$2.32 & 83.15$\pm$9.25  & 71.10$\pm$0.54 & 75.48$\pm$1.03 & 47.86$\pm$0.26 & 71.10$\pm$0.54 \\
\midrule
InfoGraph     & 76.20$\pm$1.06 & 74.44$\pm$0.31 & 72.85$\pm$1.78 & 89.01$\pm$1.13 & 70.65$\pm$1.13 & 82.50$\pm$1.42 & 53.46$\pm$1.03 & 73.03$\pm$0.87 \\
GraphCL       & 77.87$\pm$0.41 & 74.39$\pm$0.45 & 78.62$\pm$0.40 & 86.80$\pm$1.34 & 71.36$\pm$1.15 & 89.53$\pm$0.84 & 55.99$\pm$0.28 & 71.14$\pm$0.44 \\
JOAO          & 78.07$\pm$0.47 & 74.55$\pm$0.41 & 77.32$\pm$0.54 & 87.35$\pm$1.02 & 69.50$\pm$0.36 & 85.29$\pm$1.35 & 55.74$\pm$0.63 & 70.21$\pm$0.38 \\
JOAO V2       & 78.36$\pm$0.53 & 74.07$\pm$1.10 & 77.40$\pm$1.15 & 87.67$\pm$0.79 & 69.33$\pm$0.34 & 86.42$\pm$1.45 & 56.03$\pm$0.27 & 70.83$\pm$0.25 \\
SimGRACE      & 79.12$\pm$0.44 & 75.35$\pm$0.09 & 77.44$\pm$1.11 & 89.01$\pm$1.31 & 71.72$\pm$0.82 & 89.51$\pm$0.89 & 55.91$\pm$0.34 & 71.26$\pm$0.74 \\
MSSGCL        & 81.45$\pm$0.48 & 75.49$\pm$0.70 & 79.73$\pm$0.44 & 89.68$\pm$0.57 & 73.48$\pm$0.83 & 91.08$\pm$0.78 & 56.17$\pm$0.18 & 73.14$\pm$0.38 \\
CuCo          & 79.24$\pm$0.56 & 75.91$\pm$0.55 & 79.20$\pm$1.12 & 90.55$\pm$0.98 & 72.30$\pm$0.34 & 88.6$\pm$0.55  & 56.49$\pm$0.19 & 72.33$\pm$0.22 \\
\midrule
\textbf{SpecMatch-CL}  
              & \textbf{81.86}$\pm$0.36 & \textbf{76.69}$\pm$0.50 & \textbf{81.12}$\pm$1.04 & \textbf{90.87}$\pm$0.75 
              & \textbf{74.26}$\pm$0.94& \textbf{91.31}$\pm$0.79 & \textbf{57.22}$\pm$1.20 & \textbf{73.35}$\pm$0.44 \\
\bottomrule
\end{tabular}%
}
\end{table*}

\subsection{Semi-supervised training}
Table~2 shows that at the \(1\%\) label rate, SpecMatch-CL surpasses the strongest baselines on both datasets with available splits, reaching \(65.12\) on NCI1 and \(65.86\) on COLLAB, which corresponds to improvements of \(+0.49\) and \(+0.84\) points over MSSGCL, respectively. At the \(10\%\) label rate, SpecMatch-CL achieves the best results on 5 out of 6 datasets: NCI1 (\(75.67\), \(+0.81\) vs.\ Infomax/JOAOv2 at \(74.86\)), DD (\(79.21\), \(+0.32\) vs.\ MSSGCL), COLLAB (\(76.55\), \(+0.53\) vs.\ MSSGCL), RDT-B (\(91.86\), \(+1.28\) vs.\ MSSGCL), and RDT-M5K (\(55.26\), \(+0.90\) vs.\ MSSGCL), while remaining competitive on PROTEINS (\(75.06\) vs.\ \(75.76\) for MSSGCL). These results suggest that enforcing view-to-view spectral consistency is particularly beneficial when labels are limited and when preserving multi-hop neighborhood structure is critical, while maintaining strong performance in settings where existing augmentations already regularize the geometry.

\begin{table*}[t]
\centering
\caption{Results $(\%)$ on semi-supervised graph classification. "$-$" indicates that label rate is too low for the given dataset size}
\resizebox{\textwidth}{!}{%
\begin{tabular}{c|l|c c c|c c c}
\toprule
LR & Methods & NCI1 & PROTEINS & DD & COLLAB & RDT-B & RDT-M5K \\
\midrule
\multirow{10}{*}{1\%} 
& No pre-train.      & 60.72 $\pm$ 0.45 & -- & -- & 57.46 $\pm$ 0.25 & -- & -- \\
& Augmentations      & 60.49 $\pm$ 0.46 & -- & -- & 58.40 $\pm$ 0.97 & -- & -- \\
& GAE                & 61.63 $\pm$ 0.84 & -- & -- & 63.20 $\pm$ 0.67 & -- & -- \\
& Infomax            & 62.72 $\pm$ 0.65 & -- & -- & 61.70 $\pm$ 0.77 & -- & -- \\
& ContextPred        & 61.21 $\pm$ 0.77 & -- & -- & 57.60 $\pm$ 2.07 & -- & -- \\
& GraphCL            & 62.55 $\pm$ 0.86 & -- & -- & 64.57 $\pm$ 1.15 & -- & -- \\
& JOAO               & 61.97 $\pm$ 0.72 & -- & -- & 63.71 $\pm$ 0.84 & -- & -- \\
& JOAOv2             & 62.52 $\pm$ 1.16 & -- & -- & 64.51 $\pm$ 2.21 & -- & -- \\
& SimGRACE           & 64.21 $\pm$ 0.65 & -- & -- & 64.28 $\pm$ 0.98 & -- & -- \\
& MSSGCL             & 64.63 $\pm$ 0.75 & -- & -- & 65.02 $\pm$ 0.78 & -- & -- \\
& \textbf{SpecMatch-CL}   & \textbf{65.12} $\pm$ 0.65 & -- & -- & \textbf{65.86}$\pm$ 0.98 & -- & -- \\
\midrule
\multirow{10}{*}{10\%} 
& No pre-train.      & 73.72 $\pm$ 0.24 & 70.40 $\pm$ 1.54 & 73.56 $\pm$ 0.41 & 73.71 $\pm$ 0.27 & 86.63 $\pm$ 0.27 & 51.33 $\pm$ 0.44 \\
& Augmentations      & 73.59 $\pm$ 0.32 & 70.29 $\pm$ 0.64 & 74.30 $\pm$ 0.81 & 74.19 $\pm$ 0.13 & 87.74 $\pm$ 0.39 & 52.01 $\pm$ 0.20 \\
& GAE                & 74.36 $\pm$ 0.24 & 70.51 $\pm$ 0.17 & 74.54 $\pm$ 0.68 & 75.09 $\pm$ 0.19 & 87.69 $\pm$ 0.40 & 33.58 $\pm$ 0.13 \\
& Infomax            & 74.86 $\pm$ 0.26 & 72.27 $\pm$ 0.40 & 75.78 $\pm$ 0.34 & 73.76 $\pm$ 0.29 & 88.66 $\pm$ 0.95 & 53.61 $\pm$ 0.31 \\
& ContextPred        & 73.00 $\pm$ 0.30 & 70.23 $\pm$ 0.63 & 74.66 $\pm$ 0.51 & 73.69 $\pm$ 0.37 & 84.76 $\pm$ 0.52 & 51.23 $\pm$ 0.84 \\
& GraphCL            & 74.63 $\pm$ 0.25 & 74.17 $\pm$ 0.34 & 76.17 $\pm$ 1.37 & 74.23 $\pm$ 0.21 & 89.11 $\pm$ 0.19 & 52.55 $\pm$ 0.45 \\
& JOAO               & 74.48 $\pm$ 0.27 & 72.13 $\pm$ 0.92 & 75.69 $\pm$ 0.67 & 75.30 $\pm$ 0.32 & 88.14 $\pm$ 0.25 & 52.83 $\pm$ 0.54 \\
& JOAOv2             & 74.86$\pm$ 0.39 & 73.31 $\pm$ 0.48 & 75.81 $\pm$ 0.73 & 75.53 $\pm$ 0.18 & 88.79 $\pm$ 0.65 & 52.71 $\pm$ 0.28 \\
& SimGRACE           & 74.60 $\pm$ 0.41 & 74.03 $\pm$ 0.51 & 76.48 $\pm$ 0.52 & 74.74 $\pm$ 0.28 & 88.86 $\pm$ 0.62 & 53.97 $\pm$ 0.64 \\
& MSSGCL             & 74.77 $\pm$ 0.31 & \textbf{75.76} $\pm$ 0.52 & 78.89 $\pm$ 0.18 & 76.02 $\pm$ 0.13 & 90.58 $\pm$ 0.34 & 54.36 $\pm$ 0.24 \\
& \textbf{SpecMatch-CL}  & \textbf{75.67} $\pm$ 0.31 & 75.06 $\pm$ 0.68 & \textbf{79.21} $\pm$ 1.12 & \textbf{76.55} $\pm$ 0.34 & \textbf{91.86} $\pm$ 0.42 & \textbf{55.26}$\pm$ 0.35 \\
\bottomrule
\end{tabular}%
}
\end{table*}

\subsection{Transfer learning}
\begin{table}[t]
\centering
\caption{Transfer learning results (ROC-AUC \%) on benchmark datasets.}
\resizebox{\textwidth}{!}{%
\begin{tabular}{l|c|cccccccc}
\toprule
Pre-Train dataset & PPI-306K & \multicolumn{8}{c}{ZINC 2M} \\
\midrule
Pre-Train dataset & PPI & Tox21 & ToxCast & Sider & ClinTox & MUV & HIV & BBBP & Bace \\
\midrule
No Pre-Train & 64.8(1.0) & 74.6(0.4) & 61.7(0.5) & 58.2(1.7) & 58.4(6.4) & 70.7(1.8) & 75.5(0.8) & 65.7(3.3) & 72.4(3.8) \\
EdgePred     & 65.7(1.3) & 76.0(0.6) & 64.1(0.6) & 60.4(0.7) & 64.1(3.7) & 75.1(1.2) & 76.3(1.0) & 67.3(2.4) & 77.3(3.5) \\
AttrMasking  & 65.2(1.6) & 75.1(0.9) & 63.3(0.9) & 60.5(0.9) & 73.5(4.3) & 75.8(1.0) & 75.3(1.5) & 65.2(1.4) & 77.8(1.8) \\
ContextPred  & 64.4(1.3) & 73.6(0.3) & 62.6(0.6) & 59.7(1.8) & 74.0(3.4) & 72.5(1.5) & 75.6(1.0) & 70.6(1.5) & \textbf{78.8}(1.2) \\
GraphCL      & 67.88(0.85) & 75.1(0.7) & 63.0(0.4) & 59.8(1.3) & 77.5(3.8) & 76.4(0.4) & 75.1(0.7) & 67.8(2.4) & 74.6(2.1) \\
JOAO         & 64.43(1.38) & 74.8(0.6) & 62.8(0.7) & 60.4(1.5) & 66.6(3.1) & 76.6(1.7) & \textbf{76.9}(0.7) & 66.4(1.0) & 73.2(1.6) \\
SimGRACE     & 70.25(1.22) & 75.6(0.5) & 63.4(0.5) & 60.6(1.0) & 75.6(3.0) & 76.9(1.3) & 75.2(0.9) & 71.3(0.9) & 75.0(1.7) \\
\midrule
\textbf{SpecMatch-CL} & \textbf{71.75}(0.82) & \textbf{76.97}(0.45) & \textbf{64.22}(0.45) & \textbf{62.44}(1.22) & \textbf{77.78}(3.2) & \textbf{78.86}(1.37) & 76.25(0.8) & \textbf{72.88}(1.2) & 76.93(1.8) \\
\bottomrule
\end{tabular}%
}
\end{table}

Pre-training with SpecMatch-CL transfers strongly across biochemistry and biology tasks (Table 3), attaining the best ROC-AUC on 7 out of 9 downstream datasets and delivering an average gain of roughly \(+0.64\) points over the strongest baseline per dataset. Improvements are substantial on PPI (\(71.75\) vs.\ \(70.25\), \(+1.50\)), Tox21 (\(76.97\) vs.\ \(76.00\), \(+0.97\)), SIDER (\(62.44\) vs.\ \(60.60\), \(+1.84\)), MUV (\(78.86\) vs.\ \(76.90\), \(+1.96\)), and BBBP (\(72.88\) vs.\ \(71.30\), \(+1.58\)), with a more modest gain on ToxCast (\(64.22\) vs.\ \(64.10\), \(+0.12\)) and ClinTox (\(77.78\) vs.\ \(77.50\), \(+0.28\)). Performance is competitive on HIV (within \(0.65\) of the best score of \(76.9\)) and trails on BACE (\(76.93\) vs.\ \(78.8\), \(- 1.87\)), suggesting that endpoint-specific structure (e.g., motif sensitivity) may warrant tuning of the spectral loss and diffusion parameters on certain targets.

\section{Conclusion}
We presented \textbf{SpecMatch-CL}, a simple and effective loss that enforces view-to-view spectral alignment in graph contrastive learning. By matching the spectra of normalized Laplacians, the method preserves multi-scale neighborhood structure across augmentations and complements the alignment–uniformity trade-off optimized by InfoNCE. Our diffusion-kernel analysis further shows that the spectral graph-matching loss $\mathcal{L}_G$ simultaneously controls the gap to the Perfect Alignment contrastive objective and upper-bounds the Wang–Isola Uniformity loss, yielding a model-agnostic theoretical justification for the graph-matching loss. Empirically, SpecMatch-CL delivers consistent improvements on unsupervised and semi-supervised TU benchmarks and strengthens transfer performance on diverse molecular and biological datasets, all within a standard GraphCL training pipeline. Limitations include sensitivity to graph-construction choices (e.g., similarity threshold) and to the spectral graph-matching loss weight $\beta$, although we observe broad robustness across datasets in practice. Future directions include adaptive scheduling of the graph-matching loss weight, extensions to node-level and heterogeneous graphs, alternative spectral penalties (e.g., Ky–Fan norms), and multi-view generalizations that jointly align more than two augmented graphs.

\section{Acknowledgement}
The author thanks Joshua Cape for helpful comments and apologizes for listing him as a coauthor on a previous version of this article without his consent.

\bibliographystyle{unsrt}
\newpage
\bibliography{references}


\appendix

\section{Proof for Theorem 4.2}
Since the Laplacian matrices are symmetric positive semi-definite, we have $\|e^{-uL^{(1)}}\|_2 \le 1$ and $\|e^{-uL^{(2)}}\|_2 \le 1$ for all $u > 0$. We now prove that 
\begin{align*}
    \|P^{(1)} - P^{(2)}\|_F^2 \le ({t_d})^2\cdot\mathcal{L}_{G}
\end{align*}
using the Duhamel formula \cite{pazy1983semigroups}.\\
Let \( G(k) = e^{-(t_d-k)L^{(1)}} e^{-kL^{(2)}} \). We then have $G(0) = e^{-t_dL^{(1)}}$ and $G(t_d) = e^{-t_dL^{(2)}}$. Differentiate:
\begin{align*}
    &\frac{d}{dk} G(k) = e^{-(t_d-k)L^{(1)}}(L^{(1)} - L^{(2)})e^{-kL^{(2)}}\\
    \Rightarrow &e^{-t_dL^{(2)}} - e^{-t_dL^{(1)}} = \int_0^{t_d} e^{-(t_d-k)L^{(1)}}(L^{(1)} - L^{(2)})e^{-kL^{(2)}} \, dk
\end{align*}
By submultiplicativity of matrix norms we get
\begin{align*}
    \left\| P^{(1)} - P^{(2)} \right\|_F 
    &= \|e^{-{t_d}L^{(1)}} - e^{-{t_d}L^{(2)}}\|_F \le \int_0^{t_d} \|e^{-({t_d}-k)L^{(1)}}\|_2 \, \|L^{(1)} - L^{(2)}\|_F \, \|e^{-kL^{(2)}}\|_2 \, \mathrm{d}k
\end{align*}
Since $\|e^{-({t_d}-k)L^{(1)}}\|_2 \le 1$ and $\|e^{-kL^{(2)}}\|_2\le 1$, we have 
\begin{align*}
    \left\| P^{(1)} - P^{(2)} \right\|_F &\le  \|L^{(1)} - L^{(2)}\|_F \int_{0}^{t_d} 1 \, \mathrm{d}k\\
    &= {t_d} \,  \|L^{(1)} - L^{(2)}\|_F,
\end{align*}
which means 
\begin{align*}
    \left\| P^{(1)} - P^{(2)} \right\|_F^2 \le ({t_d})^2\cdot \mathcal{L}_G
\end{align*}
Because the embedding vectors are normalized, we have 
\begin{align*}
    s(z_i^{(1)}, z_i^{(2)}) = 1 - \frac{1}{2} \|z_i^{(1)} - z_i^{(2)}\|_2^2
\end{align*}

Let $z_i^{*(1)}$ and $z_i^{*(2)}$ be the embedding vectors in the Perfect Alignment case. When $z_i^{*(1)} = z_i^{*(2)}$, we have $\|z_i^{*(1)} - z_i^{*(2)}\|^2_2 = 0$.

Since all embedding vectors are unit-norm, if $z_i^{*(1)} = z_i^{*(2)}$ we have
\begin{align*}
    \left|    s(z_i^{(1)}, z_i^{(2)}) -    s(z_i^{*(1)}, z_i^{*(2)}) \right| &=  \left|1 -  \frac{1}{2}\|z_i^{(1)} - z_i^{(2)}\|^2_2 - 1 + \frac{1}{2}\|z_i^{*(1)} - z_i^{*(2)}\|^2_2 \right|\\
    &= \left|0 -  \frac{1}{2}\|z_i^{(1)} - z_i^{(2)}\|^2_2  + 0\right|\\
    &= \frac{1}{2} \|z_i^{(1)} - z_i^{(2)}\|^2_2
\end{align*}

Note that we can write the single sample contrastive loss as
\begin{align*}
    l_{i}^{(v)}
&= -\log \frac{\exp(s(z^{(v)}_i, z^{(v')}_i)/\tau)}
{\sum_{a=1}^2\sum_{k=1}^{N} \exp\!\big(s(z^{(v)}_i, z^{(a)}_k)/\tau\big) \cdot (1 - \mathbf{1}\{a=v\wedge k=i\})} \\
&= -\log \frac{\exp(s(z^{(v)}_i, z^{(v')}_i)/\tau)}
{\exp(s(z^{(v)}_i, z^{(v')}_i)/\tau) + \sum_{a=1}^2\sum_{k=1}^{N} \exp\!\big(s(z^{(v)}_i, z^{(a)}_k)/\tau\big) \cdot (1 - \mathbf{1}\{ k=i\})}\\
&= -\frac{s(z^{(v)}_i, z^{(v')}_i)}{\tau} + \log\left(\exp(s(z^{(v)}_i, z^{(v')}_i)/\tau) + \sum_{a=1}^2\sum_{k=1}^{N} \exp\!\big(s(z^{(v)}_i, z^{(a)}_k)/\tau\big) \cdot (1 - \mathbf{1}\{ k=i\})\right)
\end{align*}
Taking the derivative of $l_{i}^{(v)}$ with respect to $s(z^{(v)}_i, z^{(v')}_i)$ gives:
\begin{align*}
    \frac{\partial l_{i}^{(v)}}{\partial s(z^{(v)}_i, z^{(v')}_i)} = \frac{1}{\tau} \left( \frac{\exp(s(z^{(v)}_i, z^{(v')}_i) / \tau)}{\exp(s(z^{(v)}_i, z^{(v')}_i) / \tau) + \sum_{a=1}^2\sum_{k=1}^{N} \exp\!\big(s(z^{(v)}_i, z^{(a)}_k)/\tau\big) \cdot (1 - \mathbf{1}\{ k=i\})} - 1 \right)
\end{align*}

Note that 
\begin{align*}
    \frac{\exp(s(z^{(v)}_i, z^{(v')}_i) / \tau)}{\exp(s(z^{(v)}_i, z^{(v')}_i) / \tau) + \sum_{a=1}^2\sum_{k=1}^{N} \exp\!\big(s(z^{(v)}_i, z^{(a)}_k)/\tau\big) \cdot (1 - \mathbf{1}\{ k=i\})} \in [0,1],
\end{align*}
which means 
\[
\left| \frac{\partial l_{i}^{(v)}}{\partial s(z^{(v)}_i, z^{(v')}_i)} \right| \le \frac{1}{\tau}
\]
So $l^{(v)}_i$ is $\frac{1}{\tau}$-Lipschitz in $ s(z^{(v)}_i, z^{(v')}_i)$. If $z_i^{*(1)} = z_i^{*(2)}$ we have
\begin{align*}
    \left| l^{(v)}_i - l^{*(v)}_i \right|
&\le \frac{1}{\tau} \left|  s(z^{(v)}_i, z^{(v')}_i) -  s(z^{*(v)}_i, z^{*(v')}_i) \right|\\
& = \frac{1}{2\tau}\|z_i^{(1)} - z_i^{(2)}\|^2_2
\end{align*}
Therefore
\begin{align*}
    \left|\mathcal{L}_{C} - \mathcal{L}_{C}^{*}\right| &= \left|\sum_{v=1}^2 \sum_{i=1}^{N} l^{(v)}_i- \sum_{v=1}^2 \sum_{i=1}^{N} l^{*(v)}_i \right|\\
    &\le \sum_{v=1}^2 \sum_{i=1}^{N}  \left|l^{(v)}_i - l^{*(v)}_i\right|\\
    &\text{(by Triangle inequality)}\\
    &\le \sum_{i=1}^{N} \frac{2}{2\tau}\|z_i^{(1)} - z_i^{(2)}\|^2_2\\
    &\le \frac{({t_d})^2\ 2 c}{2\tau}  \mathcal{L}_{G}\\
    &= \frac{({t_d})^2c}{\tau} \mathcal{L}_{G}
\end{align*}
a.s. over $(z_i^{(1)}, z_i^{(2)}) \sim p_{pos}$.

\section{Proof for Theorem 4.3}
By the independent assumption
\begin{align*}
    \mathbb{E}_{L^{(1)}, L^{(2)}\sim\mathbb{P}_{L|\mathcal{G}}}[\mathcal{L}_G] &= \mathbb{E}_{L^{(1)}, L^{(2)}\sim\mathbb{P}_{L|\mathcal{G}}}\left[\|L^{(1)} - L^{(2)}\|_F^2\right]\\
    &= \mathbb{E}_{L^{(1)}, L^{(2)}\sim\mathbb{P}_{L|\mathcal{G}}}\left[\|L^{(1)} - \overline{L}  - L^{(2)} + \overline{L} \|_F^2\right]\\
    &= \mathbb{E}_{L^{(1)}, L^{(2)}\sim\mathbb{P}_{L|\mathcal{G}}}\left[\|L^{(1)} - \overline{L}\|_F^2 + \|L^{(2)} - \overline{L} \|_F^2 - 2\operatorname{tr}\left((L^{(1)} - \overline{L})^\top(L^{(2)} - \overline{L})\right)\right]\\
    &=\mathbb{E}_{L^{(1)}\sim\mathbb{P}_{L|\mathcal{G}}}\left[\|L^{(1)} - \overline{L}\|_F^2\right] + \mathbb{E}_{L^{(2)}\sim\mathbb{P}_{L|\mathcal{G}}}\left[\|L^{(2)} - \overline{L}\|_F^2\right] - \\
    &2\sum_{i,j}\mathbb{E}_{L^{(1)}, L^{(2)}\sim\mathbb{P}_{L|\mathcal{G}}}\left[(L^{(1)} - \overline{L})_{ij}(L^{(2)} - \overline{L})_{ij}\right]\\
    &= 2\mathbb{E}_{L\sim\mathbb{P}_{L|\mathcal{G}}}\left[\|L - \overline{L}\|_F^2\right] -  \\
    &2\sum_{i,j}\mathbb{E}_{L^{(1)}\sim\mathbb{P}_{L|\mathcal{G}}}\left[(L^{(1)} - \overline{L})_{ij}\right]\mathbb{E}_{L^{(2)} \sim\mathbb{P}_{L|\mathcal{G}}}\left[(L^{(2)} - \overline{L})_{ij}\right]\\
    &=2\mathbb{E}_{L\sim\mathbb{P}_{L|\mathcal{G}}}\left[\|L - \overline{L}\|_F^2\right].
\end{align*}
By Hoffman–Wielandt inequality \cite{hoffman1953variation}, we have 
\begin{align*}
    &\left(\lambda_2 - \overline{\lambda}_2\right)^2 \le \|L - \overline{L}\|_F^2,
\end{align*}
which means
\begin{align*}
    \mathbb{E}_{L\sim\mathbb{P}_{L|\mathcal{G}}}\left[(\lambda_2 - \overline{\lambda}_2)^2\right] \le \frac{1}{2}\mathbb{E}_{L^{(1)}, L^{(2)}\sim\mathbb{P}_{L|\mathcal{G}}}[\mathcal{L}_G].
\end{align*}
For any $x$,
\begin{align*}
    x^\top L x 
    &= x^\top x - x^\top D^{-1/2} A D^{-1/2} x \\
    &= \frac{1}{2} \sum_{i,j} A_{ij} \left( \frac{x_i}{\sqrt{d_i}} - \frac{x_j}{\sqrt{d_j}} \right)^2.
\end{align*}
Let $u_1 = \dfrac{D^{1/2} \mathbf{1}}{\|D^{1/2} \mathbf{1}\|}$ (the eigenvector of $L$ for eigenvalue $0$). By Rayleigh-Ritz theorem (see \cite{bhatia1997matrix}), we have 
\begin{align*}
    &\lambda_2 = \min_{\substack{x \neq 0 \\ x \perp u_1}} \frac{x^\top L x}{\|x\|_2^2}\\
    \Rightarrow\ & x^\top L x \ge \lambda_2\, \|x\|_2^2 \quad \forall x \perp u_1.
\end{align*}
Let $Z$ be the matrix of embeddings used to construct $L$. For all columns $Z_{:,r}$ of $Z$ we then define 
\begin{align*}
    &G_{:,r} := Z_{:,r} - \mu_r\textbf{1}, \\
    &x_r := D^\frac{1}{2}G_{:,r}.
\end{align*}
Notice that 
\begin{align*}
    x_r^\top u_1 &=  \frac{1}{\|D^{1/2} \mathbf{1}\|} \left(D^\frac{1}{2}(Z_{:,r} - \mu_r\textbf{1})\right)^\top D^{1/2} \mathbf{1}\\
    &= \frac{1}{\|D^{1/2}\mathbf{1}\|}\sum_{i=1}^N d_i(z_{i,r} - \mu_r)\\
    &= \frac{1}{\|D^{1/2}\mathbf{1}\|}\left(\sum_{i=1}^N d_iz_{i,r} - \mu_r \sum_{i=1}^N d_i\right) = 0,
\end{align*}
which means 
\begin{align*}
     &x_r^\top L x_r \ge \lambda_2 \|x_r\|_2^2,\\
     \Rightarrow\ &\frac{1}{2} \sum_{i,j} A_{ij} \left(G_{i,r} - G_{j,r} \right)^2 \ge \lambda_2  \sum_i d_i (G_{i,r})^2\\
     \Rightarrow\ &\frac{1}{2} \sum_{i,j} A_{ij} \left(Z_{i,r} - Z_{j,r} \right)^2 \ge \lambda_2  \sum_i d_i (Z_{i,r} - \mu_r)^2\\
      \Rightarrow\ &\frac{1}{2} \sum_{i,j} A_{ij} \|z_i - z_j\|_2^2 \ge \lambda_2  \sum_i d_i \|z_i-\mu\|_2^2\\
      \Rightarrow\ &\frac{1}{2} \sum_{i,j} \frac{A_{ij}}{\sum_k d_k}\|z_i - z_j\|_2^2 = \frac{1}{2} \sum_{i,j} \frac{A_{ij}}{\sum_{i,j} A_{ij}}\|z_i - z_j\|_2^2 \ge \lambda_2  \sum_i  \frac{d_i}{\sum_k d_k} \|z_i-\mu\|_2^2
\end{align*}
If we pick $i$ with probability $\pi_i = d_i/\sum_kd_k$ then pick $j$ with probability $P_{ij} = A_{ij}/d_i$, the joint probability of the ordered pair $(i,j)$ is 
\begin{align*}
    \mathbb{P}\{(i,j)\} = \pi_iP_{ij} = \frac{d_i}{\sum_kd_k}\cdot \frac{A_{ij}}{d_i} = \frac{A_{ij}}{\sum_kd_k} =  \frac{A_{ij}}{\sum_{i,j} A_{ij}}
\end{align*}
We also have 
\begin{align*}
    \lambda_2  \sum_i  \frac{d_i}{\sum_k d_k} \|z_i-\mu\|_2^2 &= \frac{\lambda_2}{\sum_k d_k} \sum_i d_i(\|z_i\|_2^2 + \|\mu\|_2^2 -2z_i^\top \mu)\\
    &= \lambda_2 \left(1 + \|\mu\|_2^2 - 2\mu^\top \frac{\sum_i d_iz_i}{\sum_k d_k}\right)\\
    &=  \lambda_2 (1 + \|\mu\|_2^2 - 2\|\mu\|_2^2)\\
    &=  \lambda_2 (1 - \|\mu\|_2^2),
\end{align*}
which means 
\begin{align*}
    \frac{1}{2}\ \mathbb{E}_{(i,j)\sim\pi P}\left[\|z_i - z_j\|_2^2\right] \ge  \lambda_2 (1 - \|\mu\|_2^2)
\end{align*}
However, since all embeddings are normalized, we know that $A_{ij} = 1$ if and only if $z_i^\top z_j \ge \theta$, or 
\begin{align*}
    &1 - \frac{1}{2}\|z_i - z_j \|_2^2 \ge \theta\\
    \Leftrightarrow\ &\|z_i - z_j \|_2^2 \le 2 - 2\theta.
\end{align*}
Hence, $\mathbb{E}_{(i,j)\sim\pi P}\left[\|z_i - z_j\|_2^2\right] =   \mathbb{E}_{\substack{z_i, z_j \overset{\mathrm{iid}}{\sim} p_{\text{data}}}}\left[\|z_i - z_j\|_2^2 \ \big|\  \|z_i - z_j\|_2^2 \le 2 - 2\theta\right]$. Let $p$ be the probability of $\|z_i - z_j\|_2^2 \le 2 - 2\theta$; we have
\begin{align*}
    \mathbb{E}_{\substack{z_i, z_j \overset{\mathrm{iid}}{\sim} p_{\text{data}}}}\left[\|z_i - z_j\|_2^2 \right] &= p\mathbb{E}_{\substack{z_i, z_j \overset{\mathrm{iid}}{\sim} p_{\text{data}}}}\left[\|z_i - z_j\|_2^2 \ \big|\  \|z_i - z_j\|_2^2 \le 2 - 2\theta \right] + \\
    &(1-p)\mathbb{E}_{\substack{z_i, z_j \overset{\mathrm{iid}}{\sim} p_{\text{data}}}}\left[\|z_i - z_j\|_2^2 \ \big|\  \|z_i - z_j\|_2^2 > 2 - 2\theta \right]
\end{align*}
We also have 
\begin{align*}
    &\mathbb{E}_{\substack{z_i, z_j \overset{\mathrm{iid}}{\sim} p_{\text{data}}}}\left[\|z_i - z_j\|_2^2 \ \big|\  \|z_i - z_j\|_2^2 \le 2 - 2\theta \right] \le 2 - 2\theta \\
    &\le \mathbb{E}_{\substack{z_i, z_j \overset{\mathrm{iid}}{\sim} p_{\text{data}}}}\left[\|z_i - z_j\|_2^2 \ \big|\  \|z_i - z_j\|_2^2 > 2 - 2 \theta \right],
\end{align*}
which means 
\begin{align*}
    \mathbb{E}_{\substack{z_i, z_j \overset{\mathrm{iid}}{\sim} p_{\text{data}}}}\left[\|z_i - z_j\|_2^2 \right] \ge \mathbb{E}_{\substack{z_i, z_j \overset{\mathrm{iid}}{\sim} p_{\text{data}}}}\left[\|z_i - z_j\|_2^2 \ \big|\  \|z_i - z_j\|_2^2 \le 2 - 2\theta \right].
\end{align*}
With $\|z_i - z_j\|_2^2\in[0,4]$, by the convexity of $e^{-tx}$ we have 
\begin{align*}
    e^{-t \| z_i - z_j \|_2^2} \le e^{-t\cdot 0} + \frac{e^{-t \cdot 4} - 1}{4-0}(\| z_i - z_j \|_2^2-0) =  1 - \frac{1 - e^{-4t}}{4}\| z_i - z_j \|_2^2.
\end{align*}
Therefore,
\begin{align*}
     \mathcal{L}_{\text{unif}}  &= \log \, \mathbb{E}_{\substack{z_i, z_j \overset{\mathrm{iid}}{\sim} p_{\text{data}}}}
\left[
    e^{-t \| z_i - z_j \|_2^2}
\right] \\
&\le \log \, \mathbb{E}_{\substack{z_i, z_j \overset{\mathrm{iid}}{\sim} p_{\text{data}}}}
\left[
    1 - \frac{1 - e^{-4t}}{4}\|z_i - z_j \|_2^2
\right]\\
&\le - \frac{1 - e^{-4t}}{4}\ \mathbb{E}_{\substack{z_i, z_j \overset{\mathrm{iid}}{\sim} p_{\text{data}}}}
\left[
    \|z_i - z_j \|_2^2
\right]\\
&\le - \frac{1 - e^{-4t}}{4}\mathbb{E}_{(i,j)\sim\pi P}\left[\|z_i - z_j\|_2^2\right]\\
&\le - \frac{1 - e^{-4t}}{2}\lambda_2 (1 - \|\mu\|_2^2),
\end{align*}
which means 
\begin{align*}
    \mathbb{E}_{L^{(1)}, L^{(2)}\sim\mathbb{P}_{L|\mathcal{G}}}[\mathcal{L}_{\text{unif}}] =  \mathcal{L}_{\text{unif}} &\le - \frac{1 - e^{-4t}}{2}\mathbb{E}_{L\sim\mathbb{P}_{L|\mathcal{G}}}\left[\lambda_2(1 - \|\mu\|_2^2)\right].
\end{align*}
By Cauchy–Schwarz,
\begin{align*}
    \mathbb{E}\left[\lambda_2(1 - \|\mu\|_2^2)\right] &\ge  \mathbb{E}\left[\lambda_2\right] \mathbb{E}\left[1 - \|\mu\|_2^2\right] - \sqrt{\operatorname{Var}[\lambda_2]\operatorname{Var}\left[1 - \|\mu\|_2^2\right]}.
\end{align*}
Since $(1 - \|\mu\|_2^2)\in[0,1]$, $\operatorname{Var}\left[1 - \|\mu\|_2^2\right] \le 1/4$. Moreover, 
\begin{align*}
    \mathbb{E}\left[(\lambda_2 - \overline{\lambda}_2)^2\right] &= \mathbb{E}\left[(\lambda_2 - \mathbb{E}[\lambda_2] + \mathbb{E}[\lambda_2] - \overline{\lambda}_2)^2\right] \\
    &= \mathbb{E}\left[(\lambda_2 - \mathbb{E}[\lambda_2])^2\right] + \mathbb{E}\left[(\mathbb{E}[\lambda_2] - \overline{\lambda}_2)^2\right] +2\mathbb{E}\left[(\lambda_2 - \mathbb{E}[\lambda_2])(\mathbb{E}[\lambda_2] - \overline{\lambda}_2)\right] \\
    &=  \operatorname{Var}[\lambda_2] + \mathbb{E}\left[(\mathbb{E}[\lambda_2] - \overline{\lambda}_2)^2\right],
\end{align*} 
which means 
\begin{align*}
    \operatorname{Var}[\lambda_2] \le \mathbb{E}\left[(\lambda_2 - \overline{\lambda}_2)^2\right] \le \frac{1}{2}\mathbb{E}[\mathcal{L}_G].
\end{align*}
Hence,
\begin{align*}
    \mathcal{L}_{\text{unif}} &\le - \frac{1 - e^{-4t}}{2}\mathbb{E}_{L\sim\mathbb{P}_{L|\mathcal{G}}}\left[\lambda_2\right](1 - \mathbb{E}\left[\|\mu\|_2^2\right]) + \frac{1 - e^{-4t}}{4}\sqrt{\frac{1}{2}\mathbb{E}_{L^{(1)}, L^{(2)}\sim\mathbb{P}_{L|\mathcal{G}}}[\mathcal{L}_G]}
\end{align*}
We also have
\begin{align*}
    \mathbb{E}[\lambda_2] &= \overline{\lambda}_2 + \mathbb{E}[\lambda_2 - \overline{\lambda}_2] \\
    &\ge \overline{\lambda}_2 - \mathbb{E}[|\lambda_2 - \overline{\lambda}_2|] \\
    &\ge \overline{\lambda}_2 - \sqrt{\mathbb{E}\left[(\lambda_2 - \overline{\lambda}_2)^2\right]}\quad \text{(by Cauchy-Schwarz)}\\
    &\ge \overline{\lambda}_2 - \sqrt{\frac{1}{2}\mathbb{E}_{L^{(1)}, L^{(2)}\sim\mathbb{P}_{L|\mathcal{G}}}[\mathcal{L}_G]}, 
\end{align*}which means
\begin{align*}
    \mathcal{L}_{\text{unif}} &\le - \frac{1 - e^{-4t}}{2}\left(\overline{\lambda}_2 - \sqrt{\frac{1}{2}\mathbb{E}[\mathcal{L}_G]}\right) (1 - \mathbb{E}\left[\|\mu\|_2^2\right])+  \frac{1 - e^{-4t}}{4}\sqrt{\frac{1}{2}\mathbb{E}[\mathcal{L}_G]}\\
    &= \frac{1 - e^{-4t}}{2\sqrt{2}}\left(\frac{3}{2} - \mathbb{E}\left[\|\mu\|_2^2\right]\right)\ \sqrt{\mathbb{E}_{L^{(1)}, L^{(2)}\sim\mathbb{P}_{L|\mathcal{G}}}[\mathcal{L}_G]}  - \frac{(1 - e^{-4t})}{2}\overline{\lambda}_2(1 - \mathbb{E}\left[\|\mu\|_2^2\right]).
\end{align*}

\section{Ablation study on $p$}
The ablation study results on different value of $p$ is provided in Table 4.

\begin{table*}[t]
\centering
\caption{Accuracy $(\%)$ for several value of $p$. The results indicate that SpecMatch-CL's performance is highly sensitive to the choice of $p$.}
\resizebox{\linewidth}{!}{%
\begin{tabular}{l|cccccccc}
\toprule
Value of $p$ & NCI1 & PROTEINS & DD & MUTAG & COLLAB & RDT-B & RDT-M5K & IMDB-B \\
\midrule
$p=100$ & 80.45$\pm$0.49 & 74.97$\pm$0.56 & 80.22$\pm$0.89 & 90.14$\pm$0.87 
              & 73.21$\pm$0.74& 90.71$\pm$1.07 & 56.34$\pm$0.96 & \textbf{74.75}$\pm$0.64 \\
$p=80$ & \textbf{81.86}$\pm$0.36 & \textbf{76.69}$\pm$0.50 & \textbf{81.12}$\pm$1.04 & \textbf{90.87}$\pm$0.75 
              & \textbf{74.26}$\pm$0.94& 91.31$\pm$0.79 & \textbf{57.22}$\pm$1.20 & 73.35$\pm$0.44 \\
$p=60$  & 79.32$\pm$0.58 & 75.39$\pm$0.43 & 80.45$\pm$1.12 & 89.39$\pm$0.78
              & 73.61$\pm$0.87& \textbf{92.48}$\pm$0.52 & 55.84$\pm$1.12 & 72.81$\pm$0.56 \\
\bottomrule
\end{tabular}%
}
\end{table*}

\section{Ablation study on $\beta$}
The ablation study results on different value of $\beta$ is provided in Figure 4.
\begin{figure}
    \centering
    \includegraphics[width=1\linewidth]{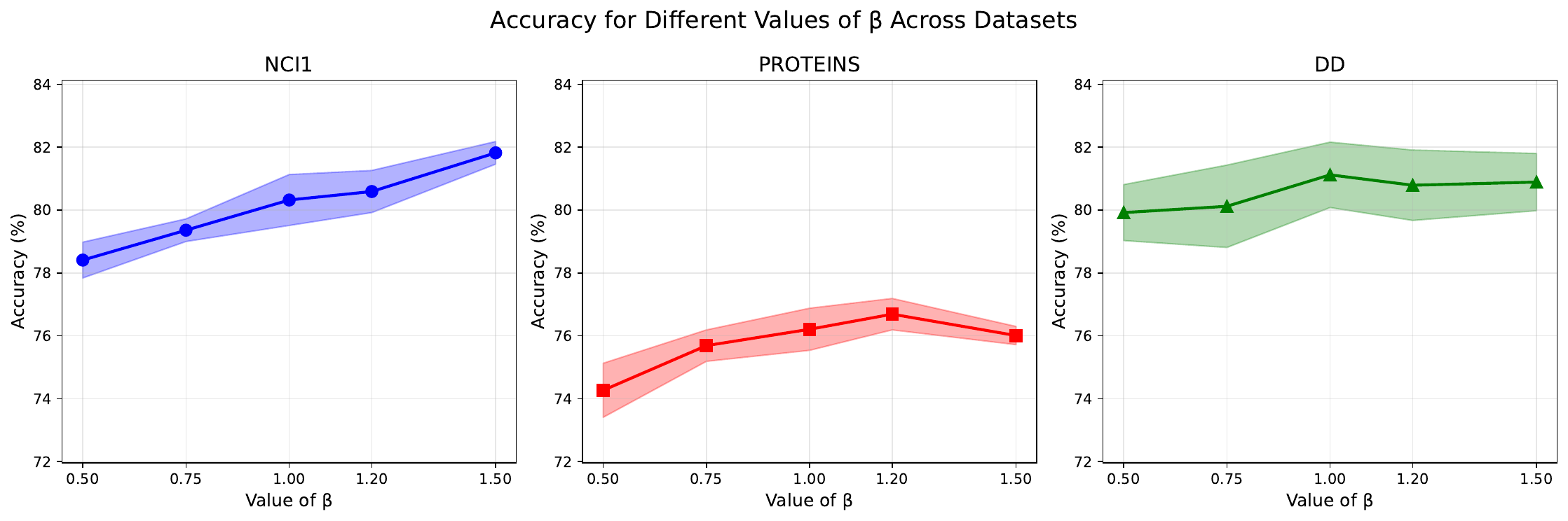}
    \caption{Ablation study for $\beta$ accross NCI1, PROTEINS and DD. As shown in the plot, $\beta$ has a significant effects on the performance of Specmatch-CL.}
    \label{fig:beta}
\end{figure}


\section{More about experiment setting}
Following GraphCL’s settings, we train with Adam optimizer a learning rate selected from \{0.01, 0.001, 0.0001\} via grid search; we use a batch size of 512 for most dataset except MUTAG (for which we use the whole datasets as a batch); and we select the number of epochs from \{20, 40, 60, 80, 100\}. All experiments are run on a single NVIDIA A100 (80GB VRAM), and each experiments take about 2-4 hours.

\section{Dataset statistics}

Dataset statistics are provided in Table 5 and 6.

\begin{table*}
\centering
\caption{Dataset statistics for unsupervised and semi-supervised experiments.}
\begin{tabular}{l|l|c|c|c}
\toprule
\textbf{Datasets} & \textbf{Category} & \textbf{Graph Num.} & \textbf{Avg. Node} & \textbf{Avg. Degree} \\
\midrule
NCI1      & Biochemical Molecules & 4110 & 29.87  & 1.08  \\
PROTEINS  & Biochemical Molecules & 1113 & 39.06  & 1.86  \\
DD        & Biochemical Molecules & 1178 & 284.32 & 715.66 \\
MUTAG     & Biochemical Molecules & 188  & 17.93  & 19.79 \\
\midrule
COLLAB    & Social Networks       & 5000 & 74.49  & 32.99 \\
RDT-B     & Social Networks       & 2000 & 429.63 & 1.15  \\
RDB-M     & Social Networks       & 2000 & 429.63 & 497.75 \\
IMDB-B    & Social Networks       & 1000 & 19.77  & 96.53 \\
\bottomrule
\end{tabular}
\end{table*}

\begin{table*}
\centering
\caption{Dataset statistics for transfer learning.}
\begin{tabular}{l|l|l|c|c|c}
\toprule
\textbf{Datasets} & \textbf{Category} & \textbf{Utilization} & \textbf{Graph Num.} & \textbf{Avg. Node} & \textbf{Avg. Degree} \\
\midrule
ZINC 2M   & Biochemical Molecules              & Pre-Training & 2{,}000{,}000 & 26.62 & 57.72 \\
PPI-306K  & Protein--Protein Interaction Nets  & Pre-Training & 306{,}925     & 39.82 & 729.62 \\
\midrule
BBBP      & Biochemical Molecules              & Finetuning   & 2{,}039       & 24.06 & 51.90 \\
ToxCast   & Biochemical Molecules              & Finetuning   & 8{,}576       & 18.78 & 38.52 \\
SIDER     & Biochemical Molecules              & Finetuning   & 1{,}427       & 33.64 & 70.71 \\
\bottomrule
\end{tabular}
\end{table*}

\end{document}